\pdfoutput=1
\documentclass{article}

\usepackage{microtype}
\usepackage{graphicx}
\usepackage{subcaption}
\usepackage{booktabs}
\usepackage{multirow}
\usepackage{array}
\usepackage{placeins}
\usepackage{hyperref}

\usepackage[preprint]{icml2026}

\makeatletter
\renewcommand{\printAffiliationsAndNotice}[1]{\global\icml@noticeprintedtrue%
  \stepcounter{@affiliationcounter}%
  {\let\thefootnote\relax\footnotetext{\hspace*{-\footnotesep}\ificmlshowauthors #1\fi%
      \forloop{@affilnum}{1}{\value{@affilnum} < \value{@affiliationcounter}}{
        \textsuperscript{\arabic{@affilnum}}\ifcsname @affilname\the@affilnum\endcsname%
          \csname @affilname\the@affilnum\endcsname%
        \else
          {\bf AUTHORERR: Missing \textbackslash{}icmlaffiliation.}
        \fi
      }.%
      \ifdefined\icmlcorrespondingauthor@text
         { }Correspondence to: \icmlcorrespondingauthor@text.
      \fi

      \ \\
      \Notice@String
    }
  }
}
\makeatother

\usepackage{amsmath}
\usepackage{amssymb}
\usepackage{mathtools}
\usepackage{amsthm}
\usepackage[capitalize,noabbrev]{cleveref}

\theoremstyle{plain}
\newtheorem{theorem}{Theorem}[section]
\newtheorem{manualtheoreminner}{Theorem}
\newenvironment{manualtheorem}[2][]{%
  \renewcommand\themanualtheoreminner{#2}%
  \if\relax\detokenize{#1}\relax\manualtheoreminner\else\manualtheoreminner[#1]\fi
}{\endmanualtheoreminner}
\newtheorem{proposition}[theorem]{Proposition}

\newtheorem{manualcorollaryinner}{Corollary}
\newenvironment{manualcorollary}[2][]{%
  \renewcommand\themanualcorollaryinner{#2}%
  \if\relax\detokenize{#1}\relax\manualcorollaryinner\else\manualcorollaryinner[#1]\fi
}{\endmanualcorollaryinner}
\theoremstyle{definition}

\theoremstyle{remark}
\newtheorem{remark}[theorem]{Remark}

\icmltitlerunning{Deconfounded Bellman Planning for Budgeted Active Feature Acquisition}

\begin{document}

\twocolumn[
  \icmltitle{Deconfounded Bellman Planning for Budgeted Active Feature Acquisition}

  \begin{icmlauthorlist}
    \icmlauthor{Yifan Wang}{cuhk}
  \end{icmlauthorlist}

  \icmlaffiliation{cuhk}{School of Medicine, The Chinese University of Hong Kong, Shenzhen, China}

  % Optional before journal/camera-ready submission:
  % \icmlcorrespondingauthor{Yifan Wang}{your_email@domain.com}

  \icmlkeywords{Active Feature Acquisition, Bellman Planning, Causal Deconfounding, Missing Not At Random, Bregman Divergence, Clinical Decision Support}

  \vskip 0.24in
]

\printAffiliationsAndNotice{}

\begin{abstract}
Active feature acquisition (AFA) selects feature groups sequentially or selectively per instance under resource budgets. Existing selectors often face a trade-off between myopic greedy rules that stall on multi-step interactions and rollout-based lookahead methods that are too slow for real-time triage. In addition, retrospective missing-not-at-random (MNAR) logs can reward the act of ordering rather than the measurement itself, while end-to-end neural selectors rarely transfer directly to frozen gradient-boosted trees (GBDTs) or shifted hospitals without target labels. We present \textbf{LIBRA} (\textbf{L}agrangian \textbf{I}nfluence \textbf{B}ellman \textbf{R}esource \textbf{A}cquisition), an amortized Bellman planning framework for budgeted AFA. LIBRA avoids propensity-ratio division under MNAR by projecting updates orthogonally via a generalized Bregman Pythagoras identity onto a within-available counterfactual centroid $\bar{p}_g(x_O)$. It computes step utilities in closed form using a quadrature-free H\'ajek--Bregman potential and trains an EMA Double-Q policy with parameter-free trapezoidal weights targeting finite-horizon acquisition AUC, paired with Lagrangian cohort allocation. Across 70 blind runs on seven public datasets, LIBRA ranks first in sequential AUC (0.7670; Holm $P \le 0.0023$) and 50\%-quota allocation (0.6597; Holm $P \le 2.91 \times 10^{-18}$) at 32.5\,ms per 1,000 samples (6.0$\times$--364.3$\times$ faster than lookahead baselines). Under controlled MNAR interventions, it keeps decoy capture at 0.00\% (vs.\ 9.1--58.7\% for baselines) and avoids observational-to-interventional sign reversal. Across 75,482 ICU stays from 77 hospitals, LIBRA cuts diagnostic blood volume by 57.4--88.9\% at matched or higher discrimination, routes directly into frozen LightGBM and XGBoost models, keeps 90\%-sensitivity safety in 34.3--43.2\% zero-draw windows, and transfers zero-shot across hospitals via unlabeled recalibration.
\end{abstract}

% =====================================================================
% SECTION 1: INTRODUCTION
% =====================================================================
\section{Introduction}
\label{sec:intro}

\begin{figure*}[t]
  \centering
  \includegraphics[width=\textwidth]{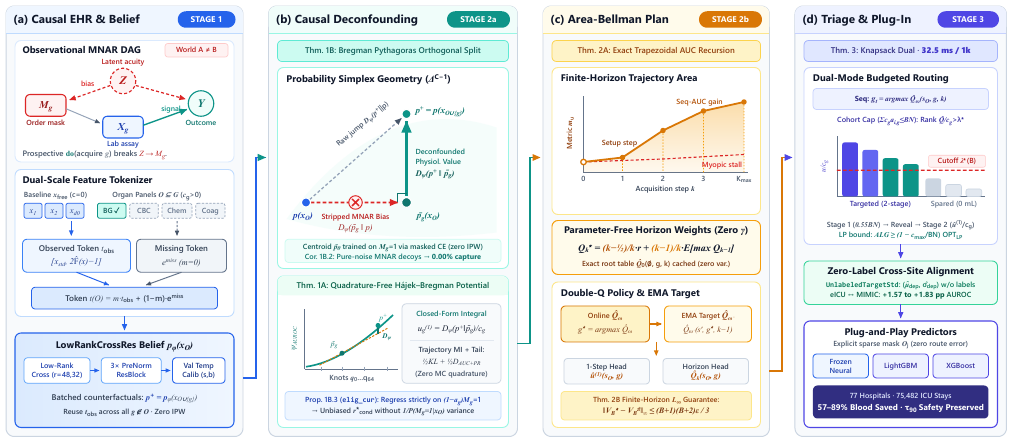}
  \vspace{-14pt}
  \caption{\textbf{Overview of LIBRA.} \textbf{(a)} Stage~1 models observational MNAR confounding ($Z \to M_g, Y$) and encodes zero-cost baselines and acquired panels using dual-scale tokens $[x_{\mathrm{std}}, 2\hat{F}(x)-1]$, learned missingness embeddings $e^{\mathrm{miss}}$, and low-rank cross-residual layers. \textbf{(b)} Stage~2a removes MNAR ordering-mask bias $D_\psi(\bar{p}_g \| p)$ orthogonally via generalized Bregman Pythagoras (Theorem~\ref{thm:mnar}) and evaluates utility in closed form with a quadrature-free H\'ajek--Bregman potential (Theorem~\ref{thm:hajek}). \textbf{(c)} Stage~2b optimizes finite-horizon trapezoidal trajectory area using parameter-free weights $(k-\frac{1}{2})/k$ and $(k-1)/k$ via EMA Double-Q recursion (Theorem~\ref{thm:bellman}). \textbf{(d)} Stage~3 performs 32.5\,ms/1k sequential or shadow-price $\lambda^\star$ cohort allocation (Theorem~\ref{thm:knapsack}), supporting unlabeled target recalibration and plug-in neural, LightGBM, and XGBoost predictors.}
  \label{fig:arch}
  \vspace{-6pt}
\end{figure*}

Active feature acquisition (AFA) decides which costly feature groups $g \in \mathcal{G}$ to observe for a given instance based on currently known covariates $x_O$, aiming to maximize predictive accuracy within per-instance or cohort budgets \citep{ma2019eddi,covert2023gdfs,gadgil2024dime,valancius2024aaco,norcliffe2025sefa}. In critical-care laboratory monitoring, for example, routine blood draws use 5.1--8.5\,mL per window and contribute to hospital-acquired anemia and transfusion demand \citep{salisbury2011blood,whitehead2019iatrogenic,halpern2014choosing}, while across-the-board rationing risks missing acute deterioration. Treating test ordering as budgeted AFA offers a natural alternative, yet current methods run into three practical obstacles (Figure~\ref{fig:arch}a).

\textbf{Myopia vs.\ lookahead latency.} Greedy one-step selectors such as GDFS \citep{covert2023gdfs} and DIME \citep{gadgil2024dime} score each candidate feature individually and stall when informative signal depends on multi-step combinations, such as a low-value context variable gating a downstream test. Lookahead methods based on partial-VAE Monte Carlo rollouts \citep[EDDI-CMI,][]{ma2019eddi}, nearest-neighbor oracle search \citep[AACO-kNN,][]{valancius2024aaco}, or latent ensembles \citep[SEFA,][]{norcliffe2025sefa} capture multi-step interactions better, but run 6$\times$--364$\times$ slower at inference. Greedy mutual information (Shannon KL) also weights probability shifts uniformly across the simplex rather than aligning rewards with cohort ranking metrics such as AUROC or precision-weighted AUPRC.

\textbf{Observational MNAR shortcuts.} Retrospective tabular datasets, especially electronic health records (EHRs), have missing-not-at-random (MNAR) recording masks $M_g \in \{0,1\}$ driven by unrecorded patient acuity $Z$ ($M_g \not\perp Y \mid X_O$) \citep{agniel2018biases,mitra2023structured,geirhos2020shortcut}. Selectors trained on raw observational posterior shifts $D(p(X_{O \cup \{g\}}) \| p(x_O))$ mix the diagnostic content of $X_g$ with the fact that a clinician ordered it ($M_g = 1$). As a result, they can score well in retrospective replay (\emph{World~A}) but reverse direction under prospective acquisition $\mathrm{do}(\text{acquire } g)$ (\emph{World~B}), since ordering a panel reveals its measurement without changing underlying disease \citep{pearl2009causality,prosperi2020causal}. Classical inverse-propensity weighting (IPW) \citep{robins1994estimation} corrects mask bias but suffers high variance when $P(M_g = 1 \mid x_O)$ is small.

\textbf{Predictor coupling and unlabeled shift.} Many AFA pipelines tie the acquisition policy to a custom masked neural network, whereas gradient-boosted decision trees (GBDTs: LightGBM, XGBoost) remain competitive on tabular data \citep{grinsztajn2022tree,ke2017lightgbm,chen2016xgboost} and are standard in clinical auditing. Deploying across hospitals also introduces covariate and unit shifts when target outcome labels are unavailable \citep{shimodaira2000covariate,lasko2024transport,pan2026transfer}.

To address these issues, we develop \textbf{LIBRA} (\emph{Lagrangian Influence Bellman Resource Acquisition}; Figure~\ref{fig:arch} and Algorithm~\ref{alg:libra}). Our contributions are:
\begin{enumerate}
  \setlength{\itemsep}{1pt}
  \setlength{\parskip}{0pt}
  \item \textbf{Quadrature-free H\'ajek--Bregman potentials} (Section~\ref{sec:hajek}): A $C^1$ piecewise-quadratic potential whose Bregman divergence matches the first-order H\'ajek influence of AUROC (alongside a tail-weighted AUPRC variant) in closed form, combined with trajectory KL telescoping for non-monotonic feature discovery.
  \item \textbf{Propensity-free causal MNAR deconfounding} (Section~\ref{sec:mnar}): Using generalized Bregman Pythagoras orthogonality around a within-available counterfactual centroid $\bar{p}_g(x_O) = \mathbb{E}[p(X_{O\cup\{g\}}) \mid x_O, M_g=1]$, LIBRA removes additive mask bias without IPW division, preserves MCAR invariance, and assigns zero reward to pure-noise MNAR decoys.
  \item \textbf{Trapezoidal Area-Bellman operator and cohort allocation} (Section~\ref{sec:bellman}): Parameter-free horizon weights $(k-\frac{1}{2})/k$ and $(k-1)/k$ optimize exact trapezoidal acquisition AUC with an $O(B^2 \varepsilon)$ error bound, learned via EMA Double-Q at 32.5\,ms/1k samples and paired with Lagrangian cohort triage and frozen GBDT routing.
  \item \textbf{Empirical validation} (Section~\ref{sec:experiments}): Across 70 blind runs on seven public benchmarks, dual-world MNAR benchmarks, and 75,482 ICU stays from 77 hospitals, LIBRA improves budget-accuracy trade-offs, eliminates decoy capture under $\mathrm{do}(\text{acquire } g)$, routes into LightGBM/XGBoost predictors, preserves $\tau_{90}$ safety in spared windows, and transfers zero-shot across hospitals.
\end{enumerate}

% =====================================================================
% SECTION 2: RELATED WORK
% =====================================================================
\section{Related Work}
\label{sec:related}

\textbf{Active feature acquisition.} Generative AFA methods model missing covariates with partial VAEs or latent diffusion to estimate expected mutual information \citep[EDDI-CMI,][]{ma2019eddi}, at substantial sampling cost during inference. Discriminative amortized policies bypass density estimation by predicting one-step conditional KL gains \citep[GDFS,][]{covert2023gdfs} or value-of-information updates \citep[DIME,][]{gadgil2024dime}, reducing runtime but plateauing on parity or gating tasks. Non-parametric nearest-neighbor search \citep[AACO-kNN,][]{valancius2024aaco} and stochastic latent encodings \citep[SEFA,][]{norcliffe2025sefa} handle multi-step lookahead better, yet depend on training-set neighbor lookups or Monte Carlo ensembles that increase latency and variance. LIBRA learns multi-step lookahead via an exact trapezoidal Bellman operator at amortized discriminative speed (32.5\,ms/1k).

\textbf{MNAR missingness and shortcut learning.} In observational EHRs, test ordering reflects clinical suspicion \citep{agniel2018biases,mitra2023structured}, creating recording shortcuts \citep{geirhos2020shortcut} that fail under prospective deployment \citep{prosperi2020causal,zhang2023active}. Unlike IPW \citep{robins1994estimation}, which divides utility estimates by propensity $\hat{P}(M_g=1 \mid x_O)$, LIBRA uses Bregman orthogonal projection onto $\bar{p}_g(x_O)$ via strictly proper scoring rules \citep{savage1971elicitation,banerjee2005clustering,gneiting2007strictly}, avoiding propensity inversion while separating conditional \texttt{elig\_cur} from propensity-attenuated regression estimands.

\textbf{Tabular predictors, calibration, and covariate shift.} Tree ensembles remain strong baselines on tabular clinical records \citep{grinsztajn2022tree,ke2017lightgbm,chen2016xgboost}. Because LIBRA outputs explicit binary panel masks independent of the final predictor, it connects directly to frozen LightGBM and XGBoost models, pairs with temperature calibration \citep{guo2017calibration} and decision-curve analysis \citep{vickers2016netbenefit}, and uses unlabeled deployment alignment \citep{shimodaira2000covariate,pan2026transfer} for zero-label site transfer.

% =====================================================================
% SECTION 3: METHOD & THEORY
% =====================================================================
\section{The LIBRA Framework and Theoretical Foundations}
\label{sec:method}

Let $X_{\mathrm{free}} \in \mathbb{R}^{d_0}$ denote zero-cost baseline covariates and $X_{\mathcal{G}} = (X_{g_1}, \dots, X_{g_G}) \in \mathbb{R}^{d_{\mathrm{pan}}}$ denote $G$ acquirable feature groups $\mathcal{G}=\{1,\dots,G\}$ with positive costs $c_g > 0$ and outcome $Y \in \{0, \dots, C-1\}$. In retrospective data $\mathcal{D}_{\mathrm{tr}} = \{(x_{i,\mathrm{free}}, x_{i,\mathcal{G}}, M_i, y_i)\}_{i=1}^{N_{\mathrm{tr}}}$, $M_{i,g} \in \{0,1\}$ records historical availability, whereas prospective deployment executes $\mathrm{do}(\text{acquire } g)$ to reveal $X_g$. For acquired subset $O \subseteq \mathcal{G}$ with binary vector $a_O \in \{0,1\}^G$ and feature-level mask $m(O) \in \{0,1\}^D$, the visible state is $s_O = (x_{\mathrm{free}}, x_O, m(O), a_O)$ and $p_\phi(x_O) := \hat{P}_\phi(Y \mid x_{\mathrm{free}}, x_O, O) \in \Delta^{C-1}$. Complete proofs of all theoretical results appear in Appendix~\ref{app:proofs}.

\subsection{Dual-Scale Cross-Residual Belief Predictor}
\label{sec:predictor}

To handle arbitrary cross-site unit scales and enable batched counterfactual evaluation across $\{O \cup \{g\}\}_{g \notin O}$, \texttt{DualScaleCrossResPredictor} embeds each continuous feature $j$ using both standardized units $x_{i,j}^{\mathrm{std}} = (x_{i,j} - \hat{\mu}_j)/\hat{\sigma}_j$ and smoothed empirical marginal CDF coordinates $x_{i,j}^{\mathrm{cdf}} = \hat{F}_j(x_{i,j}) \in [0,1]$, forming $z_{i,j} = [x_{i,j}^{\mathrm{std}},\, 2x_{i,j}^{\mathrm{cdf}} - 1]^\top \in \mathbb{R}^2$. Per-feature block-linear projections with SiLU and LayerNorm produce observed tokens $t_{i,j}^{\mathrm{obs}} \in \mathbb{R}^{d_{\mathrm{tok}}}$ ($d_{\mathrm{tok}}=12$), while unacquired features ($m_{i,j}(O)=0$) substitute learned marginal missingness tokens $e_j^{\mathrm{miss}}$:
\begin{equation}
  t_{i,j}(O) = m_{i,j}(O)\, t_{i,j}^{\mathrm{obs}} + \bigl(1 - m_{i,j}(O)\bigr)\, e_j^{\mathrm{miss}}.
\end{equation}
Concatenating $[ \mathrm{vec}(t_i(O));\, m_i(O) ]$ feeds two explicit low-rank cross-interaction layers ($r_1=48, r_2=32$),
\begin{equation}
  h_{\mathrm{cross}}(h_0, h) = h + \mathrm{LN}\!\left(h_0 \odot \bigl(V\,\mathrm{SiLU}(Uh) + b_v\bigr)\right),
\end{equation}
interleaved with $L=3$ pre-norm residual MLP blocks ($d_h=256$). Because $t_{i,j}^{\mathrm{obs}}$ is mask-independent, evaluating all $G$ counterfactual additions reuses $t_i^{\mathrm{obs}}$ in one batched pass. Validation predictions at $B=1.0$ fit a two-parameter logistic calibrator $(\hat{s}, \hat{b})$ on log-odds $\ell_i$ via L-BFGS \citep{guo2017calibration}, preserving rank discrimination while aligning probabilities.

\subsection{Quadrature-Free H\'ajek--Bregman Potentials}
\label{sec:hajek}

Rather than maximizing unweighted Shannon KL across all interior probabilities, LIBRA aligns single-step rewards with cohort evaluation metric $J$ using convex potentials $\psi : \Delta^{C-1} \to \mathbb{R}$ over $K=64$ empirical validation score quantile knots $q_0 < q_1 < \dots < q_K$ ($\Delta q_k = q_k - q_{k-1} > 0$) at outcome prevalence $\pi = \mathbb{P}(Y=1)$:
\begin{itemize}
  \setlength{\itemsep}{1pt}
  \setlength{\parskip}{0pt}
  \item \textbf{Multi-class Accuracy ($C \ge 2$):} $\psi_{\mathrm{Acc}}(p) = \max_c p_c$, yielding $D_{\psi_{\mathrm{Acc}}}^s(p^+ \| p) = \max_c p_c^+ - p_{\hat{y}(p)}^+$.
  \item \textbf{Population AUROC ($C=2$):}
  \begin{equation}
    \psi_{\mathrm{AUROC}}(p_1) = \frac{1}{\eta K} \sum_{k=1}^{K} \int_0^{p_1} \mathrm{clip}_{[0,1]}\!\Bigl(\frac{t - q_{k-1}}{\Delta q_k}\Bigr)\, \mathrm{d}t,
  \end{equation}
  where $\eta = \max(\pi(1-\pi), 0.02)$.
  \item \textbf{Tail-Weighted Binary Surrogate ($\psi_{\mathrm{AUC+PR}}$):} Replaces $1$ inside the sum with $(1 + w_k)$ using upper-tail multiplier $w_k = \frac{(q_{k-1}+q_k)/2}{\max(1 - (k-0.5)/K, 0.05)} > 0$, prioritizing high-risk precision tiers.
\end{itemize}

\begin{manualtheorem}[H\'ajek--Bregman Gain \& Closed-Form Integral]{1A}
\label{thm:hajek}
Under Bayes martingale consistency $\mathbb{E}[p(X_{O \cup \{g\}}) \mid x_O] = p(x_O)$, for any convex $\psi$ and measurable subgradient selection $s(p) \in \partial \psi(p)$, $\mathbb{E}[\psi(p^+) - \psi(p) \mid x_O] = \mathbb{E}[D_\psi^s(p^+ \| p) \mid x_O] \ge 0$. Moreover, $\psi_{\mathrm{AUROC}}$ matches the first-order Gateaux (H\'ajek) sensitivity of population AUROC locally at $F_{\mathrm{ref}}$, and for $u_k = \mathrm{clip}_{[q_{k-1}, q_k]}(p_1^+)$ and $v_k = \mathrm{clip}_{[q_{k-1}, q_k]}(p_1)$, both $\psi_{\mathrm{AUROC}}$ ($\alpha_k = (\eta K)^{-1}$) and $\psi_{\mathrm{AUC+PR}}$ ($\alpha_k = (1+w_k)(\eta K)^{-1}$) admit the exact quadrature-free divergence:
\begin{equation}
  D_\psi(p_1^+ \,\|\, p_1) = \sum_{k=1}^{K} \alpha_k\, \frac{u_k - v_k}{\Delta q_k}\!\left(p_1^+ - \frac{u_k + v_k}{2}\right) \ge 0.
\end{equation}
\end{manualtheorem}

\begin{proposition}[One-Step vs.\ Trajectory Rewards]
\label{prop:traj}
For terminal allocation (\texttt{one\_step}), LIBRA uses task-matched $D_{\psi_J}$. Along multi-step sequential trajectories (\texttt{trajectory}), LIBRA uses $\frac{1}{\log C}\mathrm{KL}(p^+ \| p)$ on public benchmarks and $0.5\,\mathrm{KL}(p^+ \| p) + 0.5\,D_{\psi_{\mathrm{AUC+PR}}}(p^+ \| p)$ on clinical cohorts. By the mutual information chain rule \citep{cover2006elements}, expected KL telescopes to $I(Y; X_{O_B} \mid X_{O_0})$, preventing zero-gain early argmax plateaus on non-monotonic parity tasks.
\end{proposition}

\subsection{Propensity-Free MNAR Orthogonal Deconfounding}
\label{sec:mnar}

When historical ordering $M_g=1$ correlates with latent acuity $Z$, $p_\phi(x_O)$ absorbs missingness cues so transitioning $m_g: 0 \to 1$ shifts $p_\phi$ even before $X_g$ conveys diagnostic information. Instead of IPW, LIBRA introduces an auxiliary centroid head $\bar{p}_\theta(x_O, g)$ trained strictly on historically measured pairs ($M_{i,g}=1, g \notin O$) by minimizing masked cross-entropy against $p_\phi(x_{i,O \cup \{g\}})$.

\begin{manualtheorem}[Bregman Pythagoras MNAR Decomposition]{1B}
\label{thm:mnar}
Let $\psi: \Delta^{C-1} \to \mathbb{R}$ be convex with fixed measurable selection $s \in \partial \psi$ (assuming $p, \bar{p}_g \in \mathrm{relint}(\Delta^{C-1})$ for $\psi_{\mathrm{KL}}$). Define the within-available counterfactual posterior centroid $\bar{p}_g(x_O) := \mathbb{E}[p(X_{O\cup\{g\}}) \mid X_O = x_O, M_g = 1]$. Whenever $\mathbb{P}(M_g=1 \mid x_O) > 0$,
\begin{align}
  \mathbb{E}\!\bigl[D_\psi^s(p^+ \,\|\, p) \mid x_O, M_g{=}1\bigr]
  &= \underbrace{\mathbb{E}\!\bigl[D_\psi^s(p^+ \,\|\, \bar{p}_g) \mid x_O, M_g{=}1\bigr]}_{\text{Deconfounded } u_{\mathrm{deconf}}(x_O, g)} \nonumber \\
  &\quad + \underbrace{D_\psi^s(\bar{p}_g \,\|\, p)}_{\text{MNAR mask bias}}.
\end{align}
\end{manualtheorem}

\begin{manualcorollary}[MCAR Invariance \& Decoy Zeroing]{1B.1}
\label{cor:mcar_decoy}
(1) Under MCAR ($M_g \perp (X_g, Y) \mid X_O$) or complete data, martingale consistency implies $\bar{p}_g(x_O) = p(x_O)$ a.s., incurring zero bias ($D_\psi^s(\bar{p}_g \| p)=0$). (2) For any pure-noise MNAR decoy ($X_g \perp Y \mid X_O, M_g=1$) whose ordering $M_g$ correlates arbitrarily with $Z$, Bayes-optimal $p(X_{O\cup\{g\}}) = \bar{p}_g(x_O)$ on $\{M_g=1\}$, forcing $u_{\mathrm{deconf}}(x_O, g) = 0$ identically while raw divergence stays strictly positive.
\end{manualcorollary}

\begin{proposition}[Propensity-Free Conditional Estimands]
\label{prop:estimands}
Because cross-entropy is strictly proper, $\bar{p}_\theta(x_O, g)$ recovers $\bar{p}_g(x_O)$ without estimating $\mathbb{P}(M_g=1 \mid x_O)$. Furthermore, regressing $D_\psi^s(p^+ \| \bar{p}_\theta)/c_g$ via MSE masked strictly to $(1 - a_g) M_g = 1$ (\texttt{elig\_cur}) recovers the exact conditional utility $r_{\mathrm{cond}}^\star(x_O, g) = \frac{1}{c_g}\mathbb{E}[D_\psi^s(p^+ \| \bar{p}_g) \mid x_O, M_g=1]$ without IPW variance blow-up, whereas including $M_g=0$ as zero targets (\texttt{unacq}) attenuates rewards by $\mathbb{P}(M_g=1 \mid x_O)$ and can invert action rankings under heterogeneous panel availability.
\end{proposition}

\subsection{Trapezoidal Area-Bellman \& Fleet Knapsack Duality}
\label{sec:bellman}

Over $k$ unit-cost steps with trajectory scores $m_0, \dots, m_k$, sequential AFA protocols evaluate the normalized trapezoidal area $\mathrm{AUC}_{\mathrm{trap}}(\pi; s_0, k) := \frac{1}{k}\sum_{u=0}^{k-1}\frac{m_u + m_{u+1}}{2} - m_0$.

\begin{manualtheorem}[Trapezoidal Area-Bellman \& $L_\infty$ Bound]{2}
\label{thm:bellman}
Under additive state-action gains $m_u - m_{u-1} = r(s_{u-1}, g_u)$ and unit cost, $\mathrm{AUC}_{\mathrm{trap}}(\pi; s_0, k) = \sum_{u=1}^{k} \frac{k - u + 0.5}{k}\, r_u$, and dynamic programming obeys $V_k^\star(s_O) = \max_{g \notin O} Q_k^\star(s_O, g)$ with $Q_0^\star \equiv 0$ and unique parameter-free operator
\begin{equation}
\begin{aligned}
  (\mathcal{T}_k f)(s_O, g) &= \frac{k{-}0.5}{k}\, r(s_O, g) \\
  &\quad + \frac{k{-}1}{k}\, \mathbb{E}\!\bigl[\max_{g' \notin O\cup\{g\}} f(s', g') \mid s_O, g\bigr].
\end{aligned}
\end{equation}
If learned $\hat{Q}_j$ satisfy $\|\hat{Q}_j - \mathcal{T}_j \hat{Q}_{j-1}\|_\infty \le \varepsilon$ for $j \in \{1,\dots,B\}$, greedy execution $\hat{\pi}_j$ satisfies $\|\hat{Q}_k - Q_k^\star\|_\infty \le \frac{k+1}{2}\varepsilon$ and $\|V_B^\star - V_B^{\hat{\pi}}\|_\infty \le \frac{(B+1)(B+2)}{3}\varepsilon$.
\end{manualtheorem}

LIBRA parameterizes a lightweight policy network $\hat{Q}_\omega(s_O, g, k)$ trained with Double-Q backups \citep{vanhasselt2016ddqn}: online $\hat{Q}_\omega$ selects $g^\star = \arg\max_{g' \notin O \cup \{g\}} \hat{Q}_\omega(s', g', k-1)$ and an EMA target $\hat{Q}_{\omega^-}$ ($\tau_{\mathrm{EMA}}=0.02$) evaluates $\hat{Q}_{\omega^-}(s', g^\star, k-1)$, while caching exact empirical root averages $\hat{Q}_0(\emptyset, g, k)$ over training samples for zero-variance step-1 selection.

\begin{manualtheorem}[Cohort Fleet Knapsack Duality]{3}
\label{thm:knapsack}
Across $N$ windows under total budget $\sum_{i,g} c_g a_{i,g} \le BN$ without conflicting per-patient varying-cost substitution constraints (unconstrained 0--1 pairs or uniform-cost top-1), ranking by $r_{i,g} = u_{i,g}/c_g$ above dual price $\lambda^\star(B)$ solves the LP relaxation up to one boundary item: $\mathrm{ALG}_{\mathrm{int}} \ge (1 - \frac{c_{\max}}{BN})\mathrm{OPT}_{\mathrm{LP}}$.
\end{manualtheorem}

\begin{remark}[MCKP Scope \& Two-Stage Clinical Heuristic]
\label{rem:mckp}
When heterogeneous costs ($c_g \in \{1.0, 1.25, 1.5\}$) pair with $\sum_g a_{i,g} \le 1$, exact LP optimality follows multiple-choice knapsack incremental slopes \citep{sinha1979mckp} (Appendix~\ref{app:proofs}). For clinical budgets $B \ge 0.75$, LIBRA uses a fast two-stage adaptive heuristic: Stage~1 spends $\le 0.55BN$ ($\le 1$ panel/window) via $\hat{Q}_\omega(s_\emptyset, g, 2)/c_g$, observes Stage-1 labs, and Stage~2 allocates the residual via $\hat{u}^{(1)}/c_g$, then passes sparse masks $O_j$ to any neural or GBDT predictor (Algorithm~\ref{alg:libra}). For cross-site transfer without target labels, \texttt{UnlabeledTargetStd} estimates $(\hat{\mu}_{\mathrm{dep}}, \hat{\sigma}_{\mathrm{dep}})$ and quantiles strictly from unlabeled target covariates.
\end{remark}

\begin{algorithm}[t]
  \caption{LIBRA Training and Budgeted Routing}
  \label{alg:libra}
  \begin{algorithmic}[1]
    \STATE {\bfseries Input:} $\mathcal{D}_{\mathrm{tr}}$, $\mathcal{D}_{\mathrm{val}}$, panels $\mathcal{G}$ with costs $\{c_g\}$, deployment $\mathcal{D}_{\mathrm{dep}}$, horizon $K_{\max}$ or cohort budget $B$.
    \STATE Train $p_\phi$ via \texttt{DualScaleCrossRes} on $O \subseteq M_i$; lock validation calibrator $(\hat{s}, \hat{b})$ and quantile knots $\{q_k\}_{k=0}^{64}$.
    \STATE Freeze $p_\phi$; fit $\bar{p}_\theta(x_O, g)$ via masked CE on $\{M_{i,g}=1\}$.
    \STATE Compute $u_{i,g}^{(1)} = D_{\psi_J}(p^+ \| \bar{p}_\theta)/c_g$ and $u_{i,g}^{(\mathrm{traj})} = D_{\mathrm{traj}}(p^+ \| \bar{p}_\theta)/c_g$ (masked by \texttt{elig\_cur} on EHRs).
    \STATE Train $\hat{Q}_\omega(s_O, g, k)$ via Double-Q MSE against $\frac{k-0.5}{k} u^{(\mathrm{traj})} + \frac{k-1}{k} \hat{Q}_{\omega^-}(s', \arg\max \hat{Q}_\omega, k-1)$; cache root $\hat{Q}_0$.
    \STATE {\bfseries Online ($\mathcal{D}_{\mathrm{dep}}$):} If label-free shift, apply \texttt{UnlabeledTargetStd}.
    \STATE \quad \emph{Sequential:} acquire $g_t = \arg\max_{g \notin O_j} \hat{Q}_\omega(s_{j,O_j}, g, k)$.
    \STATE \quad \emph{Cohort ($B$):} 2-stage $\lambda^\star$ triage (0.55$BN$ lookahead $\to$ residual $\hat{u}^{(1)}/c_g$); route $O_j$ to $p_\phi$, LightGBM, or XGBoost.
  \end{algorithmic}
\end{algorithm}

% =====================================================================
% SECTION 4: EXPERIMENTS
% =====================================================================
\section{Experiments}
\label{sec:experiments}

We organize the experimental evaluation around four questions: public benchmark accuracy and runtime (Section~\ref{sec:exp_public}), causal robustness under MNAR interventions (Section~\ref{sec:exp_mnar}), multi-hospital ICU validation, GBDT routing, and cross-site transfer (Section~\ref{sec:exp_clinical}), and mechanism ablations and safety audits (Section~\ref{sec:exp_ablation}). All runs use SHA-256 code locks and deterministic verification scripts (Appendix~\ref{app:seals}), with no patient-level files exported.

\subsection{Public Benchmarks and Runtime ($N = 70$ Blind Runs)}
\label{sec:exp_public}

\begin{table*}[t]
  \caption{Pooled 10-seed blind evaluation across seven public active-acquisition benchmarks (Seeds 5--14, $N = 70$ runs). Paired two-sided $t$-test ($t$-$P$) and Wilcoxon ($W$-$P$) $P$-values are Holm--Bonferroni adjusted across all 7 comparators.}
  \label{tab:public_macro}
  \centering
  \scriptsize
  \setlength{\tabcolsep}{3.5pt}
  \resizebox{\textwidth}{!}{\begin{tabular}{@{}lccccccccc@{}}
    \toprule
    Method & Seq-AUC & Q25 & Q50 & Q75 & Final & Latency (ms/1k) & Speedup & Seq-AUC wins (Holm $t$ / $W$) & Q50 wins (Holm $t$ / $W$) \\
    \midrule
    \textbf{LIBRA (Ours)} & \textbf{0.7670} & \textbf{0.6258} & \textbf{0.6597} & \textbf{0.6881} & \textbf{0.8625} & 32.51 & 1.00$\times$ (ref.) & --- & --- \\
    SEFA \citep{norcliffe2025sefa} & 0.7637 & 0.6109 & 0.6413 & 0.6695 & 0.8552 & 195.04 & 6.00$\times$ & 47/70 (0.0023 / 0.0040) & 68/70 ($2.9{\times}10^{-18}$ / $2.5{\times}10^{-12}$) \\
    GDFS \citep{covert2023gdfs} & 0.7565 & 0.5858 & 0.5945 & 0.6102 & 0.8461 & 17.34 & 0.53$\times$ & 57/70 ($3.2{\times}10^{-8}$ / $2.9{\times}10^{-9}$) & 70/70 ($2.6{\times}10^{-21}$ / $2.5{\times}10^{-12}$) \\
    LIBRA-SamePred & 0.7543 & 0.6152 & 0.6467 & 0.6744 & 0.8486 & 20.55 & 0.63$\times$ & 64/70 ($2.9{\times}10^{-10}$ / $5.3{\times}10^{-11}$) & 61/70 ($4.7{\times}10^{-14}$ / $1.0{\times}10^{-10}$) \\
    AACO-kNN \citep{valancius2024aaco} & 0.7301 & 0.6126 & 0.6430 & 0.6708 & 0.8175 & 4,096.18 & 126.0$\times$ & 62/70 ($3.3{\times}10^{-7}$ / $3.3{\times}10^{-11}$) & 64/70 ($2.3{\times}10^{-17}$ / $1.2{\times}10^{-11}$) \\
    DIME \citep{gadgil2024dime} & 0.7249 & 0.5960 & 0.6137 & 0.6318 & 0.7775 & 20.38 & 0.63$\times$ & 59/70 ($7.0{\times}10^{-5}$ / $2.0{\times}10^{-9}$) & 69/70 ($9.6{\times}10^{-21}$ / $2.5{\times}10^{-12}$) \\
    FixedMLP & 0.7195 & 0.6063 & 0.6329 & 0.6562 & 0.7828 & 16.21 & 0.50$\times$ & 61/70 ($2.6{\times}10^{-6}$ / $2.9{\times}10^{-9}$) & 68/70 ($3.1{\times}10^{-19}$ / $4.3{\times}10^{-12}$) \\
    EDDI-CMI \citep{ma2019eddi} & 0.7174 & 0.6012 & 0.6266 & 0.6514 & 0.8143 & 11,842.71 & 364.3$\times$ & 68/70 ($2.8{\times}10^{-11}$ / $2.8{\times}10^{-12}$) & 65/70 ($1.8{\times}10^{-17}$ / $4.8{\times}10^{-12}$) \\
    \bottomrule
  \end{tabular}}
  \vspace{-4pt}
\end{table*}

\begin{figure*}[t]
  \centering
  \includegraphics[width=\textwidth]{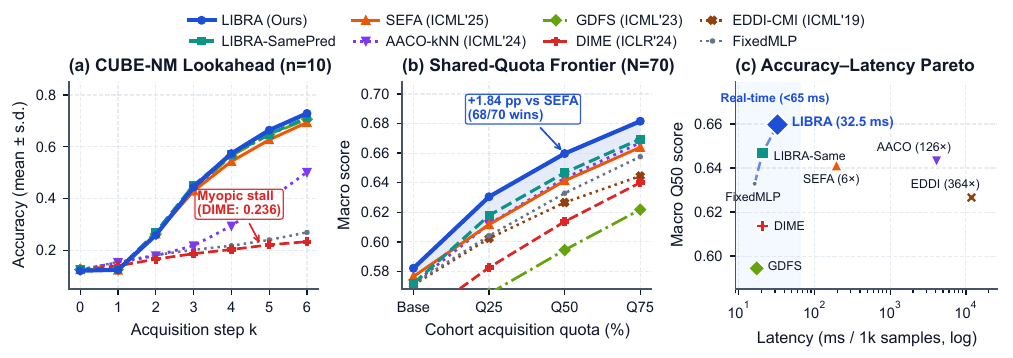}
  \vspace{-14pt}
  \caption{\textbf{Public blind benchmarks ($N=70$ runs: 7 datasets $\times$ Seeds 5--14).} \textbf{(a)} Non-monotonic CUBE-NM sequential accuracy ($n=10$ seeds; shaded band $\pm 1$ s.d. for LIBRA/SEFA). \textbf{(b)} Shared-quota macro score from Base through Q25, Q50, Q75 ($N=70$). \textbf{(c)} Macro Q50 vs.\ log-scale inference latency (ms/1k), placing LIBRA on the real-time Pareto frontier ($<65$\,ms).}
  \label{fig:public}
  \vspace{-6pt}
\end{figure*}

Following validation selection on development Seeds~0--2 (confirming EMA Double-Q: 0.7244 vs.\ 0.7237 Seq-AUC; Appendix~Table~\ref{tab:app_val}), we evaluated two disjoint blind test cohorts (Seeds~5--9 and confirmatory Seeds~10--14) across seven benchmarks (Syn1, CUBE-NM, Housing, Bank, MiniBooNE, PhysioNet, Diabetes; $N=70$ runs) against six external baselines and \texttt{LIBRA-SamePred}, which freezes the exact \texttt{FixedMLP} weights with zero base-step parity error.

As shown in Table~\ref{tab:public_macro} and Figure~\ref{fig:public}, LIBRA ranks first on all five macro metrics: Seq-AUC (0.7670), Q25 (0.6258), Q50 (0.6597), Q75 (0.6881), and Final (0.8625). After Holm--Bonferroni correction, LIBRA outperforms all seven comparators on both co-primary endpoints, including SEFA (Seq-AUC: 47/70 wins, $P_{\mathrm{Holm}}=0.0023$; Q50: 68/70 wins, $P_{\mathrm{Holm}}=2.9\times 10^{-18}$). On non-monotonic CUBE-NM (Figure~\ref{fig:public}a), myopic DIME and FixedMLP stall below 0.27, whereas LIBRA selects the zero-gain setup coordinate at step~1 and reaches 0.729 terminal accuracy. Notably, \texttt{LIBRA-SamePred} raises \texttt{FixedMLP} from 0.7195 to 0.7543 Seq-AUC ($+3.48$\,pp) and 0.6467 Q50, beating every external baseline solely through planning, while LIBRA runs in 32.51\,ms/1k (6.0$\times$ faster than SEFA, 126.0$\times$ than AACO-kNN, and 364.3$\times$ than EDDI-CMI; Figure~\ref{fig:public}c). Full per-dataset curves and standalone batch tests (where Seeds~10--14 Q50 wins 33/35 vs.\ SEFA, $P_{\mathrm{Holm}}=1.03\times 10^{-8}$, while standalone Seq-AUC over SEFA is 0.7656 vs.\ 0.7634, $P=0.1405$) are reported in Appendix~\ref{app:public}.

\subsection{Causal Robustness Under Dual-World MNAR Interventions}
\label{sec:exp_mnar}

\begin{figure*}[t]
  \centering
  \includegraphics[width=\textwidth]{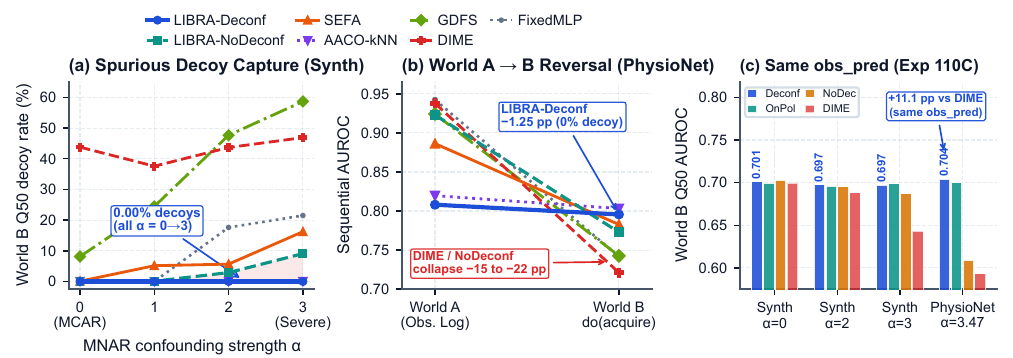}
  \vspace{-14pt}
  \caption{\textbf{Dual-world causal MNAR interventions (Exps.~108 \& 110C).} \textbf{(a)} World~B Q50 decoy capture (\%) across confounding $\alpha \in \{0,1,2,3\}$ ($n=5$ seeds); LIBRA-Deconf stays at 0.00\%. \textbf{(b)} Observational World~A to interventional World~B ($\mathrm{do}(\text{acquire})$) sign reversal on semi-synthetic PhysioNet ($\alpha=3.47$). \textbf{(c)} Identical frozen \texttt{obs\_pred} forensics ($\Delta B_0 = 0.0$): deconfounding alone drives $+11.1$\,pp World~B Q50 gain over DIME on PhysioNet and eliminates Synth $\alpha=3.0$ decoys.}
  \label{fig:mnar}
  \vspace{-6pt}
\end{figure*}

To test Theorem~\ref{thm:mnar} directly, we evaluate Synthetic EHR ($\alpha \in \{0,1,2,3\}$) and Semi-Synthetic PhysioNet ICU ($\alpha = 3.47$) benchmarks pairing informative panels with pure-noise decoys ordered via $P(M_{\mathrm{spur}}=1 \mid Z) = \sigma(\alpha Z + \tau)$ across \emph{World~A} (observational replay) and \emph{World~B} ($\mathrm{do}(\text{acquire } g)$). At $\alpha = 0$ (MCAR), \texttt{LIBRA-Deconf} and \texttt{LIBRA-NoDeconf} match closely (0.8084 vs.\ 0.8085 terminal AUROC, $P=0.621$), consistent with Corollary~\ref{cor:mcar_decoy}. As confounding increases to $\alpha = 3.0$, World~B Q50 decoy capture rises to 58.7\% (GDFS), 46.9\% (DIME), 16.2\% (SEFA), and 9.1\% (NoDeconf), whereas \texttt{LIBRA-Deconf} stays at 0.00\% across all $\alpha$ (Figure~\ref{fig:mnar}a) and reaches 0.8101 terminal AUROC ($+3.95$\,pp over NoDeconf, $P=1.85\times 10^{-4}$; $+13.82$\,pp over DIME, $P=8.7\times 10^{-5}$).

On PhysioNet (Figure~\ref{fig:mnar}b), un-deconfounded models show clear World~A to World~B sign reversal: FixedMLP (0.9425), DIME (0.9378), GDFS (0.9241), and NoDeconf (0.9229) score highest under observational masks in World~A, then drop by $-15.00$ to $-21.66$\,pp in World~B (96--100\% Q50 decoys). By contrast, \texttt{LIBRA-Deconf} changes by only $-1.25$\,pp, keeps Q50 decoys at 0.00\%, and achieves the highest World~B Q50 AUROC (0.6887 vs.\ NoDeconf 0.6062, $P=8.45\times 10^{-4}$). Holding the Stage-1 predictor fixed to identical weights (\texttt{obs\_pred}, Exp.~110C; Figure~\ref{fig:mnar}c) isolates the effect of Stage-2 $\bar{p}_g$ deconfounding: it restores 100.0\% step-1 diagnostic targeting on both benchmarks, raises Synth $\alpha=3.0$ World~B terminal AUROC from 0.7285 (DIME) / 0.7640 (NoDeconf) to 0.8100 ($P \le 0.00031$), and improves PhysioNet Q50 by $+11.10$\,pp over DIME (0.7043 vs.\ 0.5933).

\subsection{Multi-Hospital ICU Validation, GBDT Routing, and Transfer}
\label{sec:exp_clinical}

\begin{figure*}[t]
  \centering
  \includegraphics[width=\textwidth]{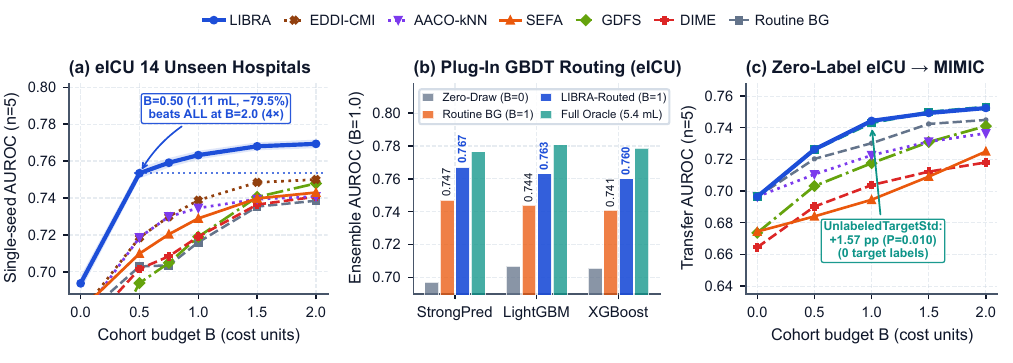}
  \vspace{-14pt}
  \caption{\textbf{Multi-center ICU validation, plug-in GBDT routing, and zero-label transfer.} \textbf{(a)} Single-seed AUROC ($n=5$) across budgets $B \in [0, 2.0]$ on 14 completely unseen external eICU hospitals ($N=15,374$ windows); LIBRA at $B=0.50$ (1.11\,mL, $-79.5\%$ blood) outperforms every baseline at $B=2.00$. \textbf{(b)} Lossless mask transfer at $B=1.00$ to frozen LightGBM and XGBoost on eICU. \textbf{(c)} Zero-label \texttt{eICU $\to$ MIMIC} transfer and \texttt{UnlabeledTargetStd} recalibration gain ($+1.57$\,pp AUROC, $P=0.010$).}
  \label{fig:clinical}
  \vspace{-6pt}
\end{figure*}

\begin{table*}[t]
  \caption{Multi-backbone GBDT routing, calibration, and validation-locked $\tau_{90}$ safety audit at $B=1.00$ across four ICU cohorts.}
  \label{tab:clinical_gbdt}
  \centering
  \scriptsize
  \setlength{\tabcolsep}{3.2pt}
  \resizebox{\textwidth}{!}{\begin{tabular}{@{}llcccccccc@{}}
    \toprule
    Cohort (Test windows; factual draw) & Configuration ($B=1.0$ unless noted) & Ens.\ AUROC & Ens.\ AUPRC & ECE $\downarrow$ & $\tau_{90}$ Spec $\uparrow$ & Spared 0-mL & Spared Miss/1k (Oracle) & Acq.\ Miss/1k (Rout.\ BG) & Blood (Saved \%) \\
    \midrule
    \textbf{eICU} ($15,374$, 14 hosp.; 5.43\,mL) & LGBM\_B0 / Routine\_BG & 0.7070 / 0.7439 & 0.5189 / 0.5865 & 0.070 / 0.054 & 27.0 / 28.3\% & 100 / 0\% & --- / 36.74 & --- / 27.57 & 0.0 / 2.09\,mL \\
     & \textbf{StrongPred\_LIBRA / LGBM\_Routed} & \textbf{0.7672} / 0.7631 & \textbf{0.6015} / 0.6002 & 0.051 / \textbf{0.050} & 36.9 / \textbf{37.1\%} & \textbf{36.9\%} & \textbf{48.55} (51.06) & \textbf{19.38} (27.57) & 2.27\,mL (\textbf{58.2\%}) \\
    \midrule
    \textbf{MIMIC-IV} ($6,464$; 7.51\,mL) & LGBM\_B0 / Strong\_Routine\_BG & 0.7671 / 0.7885 & 0.6783 / 0.7210 & \textbf{0.013} / 0.029 & 33.7 / 32.5\% & 100 / 0\% & --- / 32.87 & --- / 33.49 & 0.0 / 1.84\,mL \\
     & \textbf{StrongPred\_LIBRA / LGBM\_Routed} & \textbf{0.8082} / 0.7995 & \textbf{0.7346} / 0.7244 & 0.023 / \textbf{0.016} & \textbf{41.3} / 38.2\% & \textbf{34.3\%} & \textbf{55.52} (57.74) & \textbf{17.54} (33.49) & 1.95\,mL (\textbf{74.0\%}) \\
    \midrule
    \textbf{Fuwai Decomp.} ($45,781$, '23; 8.52\,mL) & LGBM\_B0 / FullOracle & 0.7631 / 0.7786 & 0.5306 / 0.5500 & \textbf{0.108} / 0.112 & 36.6 / 37.7\% & 100 / 0\% & --- / 39.28 & --- / 5.30 & 0.0 / 8.52\,mL \\
     & \textbf{StrongPred\_LIBRA / LGBM\_Routed} & \textbf{0.7676} / \textbf{0.7676} & 0.5277 / \textbf{0.5363} & 0.112 / 0.110 & 35.6 / 36.4\% & \textbf{43.2\%} & 41.02 (39.28) & 6.36 (5.30) & 1.67\,mL (\textbf{80.4\%}) \\
    \midrule
    \textbf{Fuwai AKI} ($2,141$, low-headroom; 5.11\,mL) & \textbf{LGBM\_B0} / StrongPred\_LIBRA & \textbf{0.8240} / 0.8217 & 0.4914 / 0.4757 & \textbf{0.051} / 0.063 & 52.7 / 44.0\% & 100 / 35.7\% & --- / 27.81 & --- / \textbf{10.98} & 0.0 / 1.79\,mL \\
     & \textbf{XGB\_LIBRA\_Routed} & 0.8222 & \textbf{0.5043} & 0.052 & \textbf{53.2\%} & \textbf{35.7\%} & --- & --- & 1.79\,mL (65.0\%) \\
    \bottomrule
  \end{tabular}}
  \vspace{-4pt}
\end{table*}

We evaluated LIBRA across 75,482 surgical and ICU stays (208,686 total windows; 69,760 held-out test windows) spanning 77 hospitals in the US \citep[\textbf{eICU}, 61 dev $\to$ 14 unseen external hospitals; \textbf{MIMIC-IV CVICU};][]{pollard2018eicu,johnson2023mimic} and China (\textbf{Fuwai Decompensation} and \textbf{Fuwai Incident 6--72\,h KDIGO AKI} \citep{khwaja2012kdigo}, 2022 dev $\to$ 2023 prospective holdout; Appendix~Table~\ref{tab:app_cohorts}). All nine comparators used the same validation temperature calibration.

\textbf{Discrimination and blood-volume compression.} At $B=1.00$ (57.4--80.4\% blood reduction), LIBRA ranks first in single-seed AUROC and AUPRC across all four cohorts (20/20 runs; eICU: $0.7632\pm0.0023$ vs.\ EDDI $0.7388$, SEFA $0.7289$; MIMIC: $0.8059\pm0.0025$ vs.\ GDFS $0.7861$; Appendix~Table~\ref{tab:app_clin9}). On the 14 unseen eICU hospitals, LIBRA's 5-seed ensemble leads in AUROC (0.7686), AUPRC (0.6030), and Brier score (0.1572; patient bootstrap $P < 0.001$ and 14-hospital-cluster bootstrap $P \le 0.050$ vs.\ all 8 baselines), scoring 15,374 windows in 36.8\,ms (8.2$\times$--15.6$\times$ faster than SEFA/AACO/EDDI). Even at $B=0.50$ (1.11\,mL in eICU, $-79.5\%$; 0.83\,mL in MIMIC, $-88.9\%$), LIBRA outperforms all competing policies running at four times the budget ($B=2.00$; Figure~\ref{fig:clinical}a, Appendix~Table~\ref{tab:app_iso}).

\textbf{Plug-in GBDT routing and $\tau_{90}$ safety.} LIBRA's sparse binary masks transfer without parity loss into frozen LightGBM and XGBoost models (Figure~\ref{fig:clinical}b, Table~\ref{tab:clinical_gbdt}): on eICU, \texttt{LGBM\_LIBRA\_Routed} improves from 0.7439 (Routine\_BG) to 0.7631 AUROC (ECE 0.0495); on MIMIC-IV from 0.7885 to 0.7995; and on Fuwai Decompensation GBDT AUPRC reaches 0.5363. Under validation-locked 90\%-sensitivity thresholds ($\tau_{90}$), LIBRA spares 34.3--43.2\% of windows from any blood draw (0\,mL), with event rates 2.0$\times$--3.4$\times$ lower than acquired windows. Spared-window missed events match or slightly improve on the unbudgeted FullOracle ceiling (eICU: 48.55 vs.\ 51.06/1k; MIMIC: 55.52 vs.\ 57.74/1k), while acquired-stratum missed crises fall by 47.6\% (MIMIC) and 29.7\% (eICU) relative to Routine BloodGas alongside an 8.0--8.8\,pp gain in overall specificity (Appendix~Figure~\ref{fig:app_clin}b).

\textbf{Zero-label cross-hospital transfer.} Across 122 harmonized variables between single-center MIMIC-IV and 75-hospital eICU with no target outcome labels (Figure~\ref{fig:clinical}c, Appendix~Table~\ref{tab:app_xfer}), LIBRA leads in both directions ($B=1.00$ ensemble AUROC 0.6793 MIMIC$\to$eICU; 0.7494 eICU$\to$MIMIC, patient $P \le 0.008$). Adding \texttt{UnlabeledTargetStd} yields $+1.57$ to $+1.83$\,pp AUROC gains on eICU$\to$MIMIC (raw $P \le 0.0104$; Brier $0.221 \to 0.202$) and $+1.44$\,pp for end-to-end LIBRA on MIMIC$\to$eICU ($P=0.0098$).

\subsection{Ablations, Saturation Boundaries, and Multiplicity}
\label{sec:exp_ablation}

\begin{table}[t]
  \caption{Component ablations on public Seq-AUC ($n=5$, Exp.~105) and clinical AUPRC at $B=1.00$ ($n=5$, Exp.~106).}
  \label{tab:ablation}
  \centering
  \scriptsize
  \setlength{\tabcolsep}{3.0pt}
  \resizebox{\columnwidth}{!}{\begin{tabular}{@{}lcccc@{}}
    \toprule
    Variant & Syn1 (AUC) & CUBE-NM & eICU (PR) & Fuwai AKI \\
    \midrule
    \textbf{LIBRA (Full)} & \textbf{0.7771} & \textbf{0.4182} & 0.5950 & 0.4681 \\
    w/o Trapezoidal Bellman & 0.7500 & 0.3584 & 0.5948 & 0.4642 \\
    w/o H\'ajek Influence & 0.7769 & 0.4212 & 0.5925 & 0.4674 \\
    w/o LowRankCross & 0.7765 & 0.4191 & 0.5942 & 0.4494 \\
    w/o MNAR Centroid & 0.7771 & 0.4182 & 0.5968 & 0.4764 \\
    \bottomrule
  \end{tabular}}
  \vspace{-6pt}
\end{table}

Table~\ref{tab:ablation} separates the contribution of each module: removing trapezoidal Bellman lookahead drops CUBE-NM by $-5.98$\,pp and Syn1 by $-2.71$\,pp; removing low-rank cross layers reduces Fuwai AKI AUPRC by $-1.87$\,pp; and removing H\'ajek influence lowers eICU AUPRC by $-0.25$\,pp. Dropping MNAR deconfounding appears slightly higher in retrospective eICU/AKI logs because World~A rewards ordering shortcuts (Section~\ref{sec:exp_mnar}), underscoring the need for interventional World~B testing.

Three remaining audits characterize where LIBRA helps and where it saturates: (1)~\textbf{Saturation boundary:} On Fuwai Incident AKI, 152 intraoperative covariates already saturate baseline discrimination, so zero-draw \texttt{LGBM\_B0} reaches 0.8240 AUROC vs.\ FullOracle 0.8299 ($P=0.654$ vs.\ LIBRA 0.8217); comparing $B=0$ GBDTs against FullOracle reliably flags low-headroom regimes where extra testing adds little. (2)~\textbf{Ordering behavior, cost invariance, and subgroup equity:} Without hard-coded heuristics, LIBRA spares 53--57\% of low-risk windows, favors blood gases at 0--6\,h and chemistry/CBC at 24--72\,h, adds $+9.29$\,pp AUROC when prior lactate is stale ($>6$\,h; Appendix~Figure~\ref{fig:app_clin}c), adapts panel shares across Clinical/\allowbreak Uniform/\allowbreak mL cost tables within $\le 0.40$\,pp AUROC, and improves 82.4\% (112/136) of evaluations across 34 demographic, surgical, and hospital-tier subgroups. (3)~\textbf{Multiplicity:} Across 326 pre-specified tests in 18 families, 185/326 (56.7\%) remain significant at family-wise Holm $P<0.05$ (100\% of pooled public tests; Appendix~Figure~\ref{fig:app_clin}d).

% =====================================================================
% SECTION 5: CONCLUSION AND LIMITATIONS
% =====================================================================
\section{Conclusion and Limitations}
\label{sec:conclusion}

We introduced LIBRA, a budgeted AFA method combining closed-form H\'ajek--Bregman utility rewards, propensity-free Bregman Pythagoras MNAR deconfounding, and parameter-free trapezoidal Double-Q Bellman planning. Across public benchmarks, synthetic/semi-synthetic interventions, and multi-hospital ICU cohorts, it runs at 32.5\,ms/1k samples, avoids MNAR decoy shortcuts, transfers sparse acquisition masks directly to GBDT predictors, and adapts across sites without target labels.

\textbf{Limitations.} First, while Theorem~\ref{thm:mnar} removes additive ordering-mask jumps and assigns zero utility to pure-noise decoys, informative tests whose treatment-effect magnitude varies with unobserved severity $Z$ remain conditioned on $\{M_g=1\}$ absent instrumental or randomized exploration. Second, retrospective hospital holdouts evaluate World~A observed masks; silent-mode and cluster-randomized stewardship trials are required before clinical deployment. Third, \texttt{UnlabeledTargetStd} addresses $P(X)$ and unit shifts but not severe label shift or concept drift $P(Y \mid X)$ without periodic recalibration, and the two-stage clinical router approximates exact multiple-choice knapsack allocation under heterogeneous panel costs.

% =====================================================================
% IMPACT STATEMENT (EXEMPT FROM 8-PAGE LIMIT; MUST START ON PAGE 9+)
% =====================================================================
\clearpage
\section*{Impact Statement}

This work studies cost-sensitive active feature acquisition with application to clinical laboratory stewardship in intensive care. By sparing unnecessary diagnostic blood draws in low-risk windows (34.3--43.2\% zero-draw windows at $\tau_{90}$ safety parity) while concentrating multi-organ panels on unstable or stale-lactate patients, budgeted AFA could help reduce iatrogenic phlebotomy blood loss (57.4--88.9\% in our simulations), hospital-acquired anemia, transfusion demand, and testing costs without delaying crisis detection. All clinical analyses in this study were run under read-only governance on de-identified retrospective and temporal-prospective datasets (MIMIC-IV, eICU, and IRB-approved Fuwai Hospital records) with no patient-level data exported. Any future bedside use would require prospective silent-mode auditing, local $\tau_{90}$ safety checks, and clinician oversight to monitor for unmeasured acuity confounding, concept drift, and subgroup performance gaps.

% =====================================================================
% REFERENCES
% =====================================================================
\bibliography{libra_icml}
\bibliographystyle{icml2026}

% =====================================================================
% APPENDICES
% =====================================================================
\newpage
\appendix
\onecolumn
\renewcommand{\thefigure}{A\arabic{figure}}
\setcounter{figure}{0}
\renewcommand{\thetable}{A\arabic{table}}
\setcounter{table}{0}

\section{Complete Theoretical Proofs and Scope Analysis}
\label{app:proofs}

\subsection{Proof of Theorem~\ref{thm:hajek} (H\'ajek--Bregman Gain \& Closed-Form Integral)}
\begin{proof}
\textbf{(1) Proper Bregman Gain Identity.} By definition of the generalized Bregman divergence with measurable subgradient selection $s(p) \in \partial \psi(p)$,
\[
  D_\psi^s(p^+ \,\|\, p) = \psi(p^+) - \psi(p) - \langle s(p),\, p^+ - p \rangle \ge 0.
\]
Setting $p^+ = p(X_{O\cup\{g\}})$ and $p = p(x_O)$ and taking conditional expectations given $X_O = x_O$, Bayes martingale consistency $\mathbb{E}[p^+ \mid x_O] = p(x_O)$ eliminates the linear inner product because $s(p(x_O))$ is $x_O$-measurable: $\mathbb{E}[\langle s(p), p^+ - p \rangle \mid x_O] = \langle s(p), 0 \rangle = 0$.

\textbf{(2) First-Order H\'ajek Sensitivity.} For binary scores $p_1$ under continuous conditionals $F_0, F_1$, prevalence $\pi$, and mixture $F_{\mathrm{ref}} = \pi F_1 + (1-\pi)F_0$, $J_{\mathrm{AUROC}}(P) = \mathbb{P}_{X\sim P_1, X'\sim P_0}(p_1(X) > p_1(X'))$ has Gateaux influence derivative $h'(p_1) = \frac{F_{\mathrm{ref}}(p_1) - \pi}{\pi(1-\pi)}$ around $P_{\mathrm{ref}}$. Because subtracting the affine term $\frac{\pi}{\pi(1-\pi)}p_1$ leaves any Bregman divergence invariant, $\psi'_{\mathrm{AUROC}}(p_1) = \frac{\hat{F}_{\mathrm{ref}}(p_1)}{\max(\pi(1-\pi), 0.02)}$ matches the first-order H\'ajek derivative across the empirical quantile knots $\{q_k\}_{k=0}^K$ ($K=64$).

\textbf{(3) Closed-Form Integral.} Note $\psi'_{\mathrm{AUROC}}(t) = \sum_{k=1}^K \alpha_k\, \mathrm{clip}_{[0,1]}\bigl(\frac{t - q_{k-1}}{\Delta q_k}\bigr)$ is piecewise linear and continuous, so $\psi_{\mathrm{AUROC}}$ (and $\psi_{\mathrm{AUC+PR}}$) is convex and globally $C^1$. Integrating $D_\psi(p_1^+ \| p_1) = \int_{p_1}^{p_1^+} [\psi'(p_1^+) - \psi'(t)]\,\mathrm{d}t$ term-by-term over intervals $[v_k, u_k]$ with $u_k = \mathrm{clip}_{[q_{k-1}, q_k]}(p_1^+)$ and $v_k = \mathrm{clip}_{[q_{k-1}, q_k]}(p_1)$ gives
\[
  D_\psi(p_1^+ \,\|\, p_1) = \sum_{k=1}^K \alpha_k \int_{v_k}^{u_k} (p_1^+ - t)\,\frac{\mathrm{d}t}{\Delta q_k} = \sum_{k=1}^K \alpha_k\,\frac{u_k - v_k}{\Delta q_k}\!\left(p_1^+ - \frac{u_k + v_k}{2}\right) \ge 0.
\]
\end{proof}

\subsection{Proof of Proposition~\ref{prop:traj}}
Along any sequence $O_0 \subset O_1 \subset \dots \subset O_B$, the mutual information chain rule \citep{cover2006elements} gives $\sum_{u=1}^B \mathbb{E}[\mathrm{KL}(p(X_{O_u}) \| p(X_{O_{u-1}}))] = I(Y; X_{O_B} \mid X_{O_0})$. Thus maximizing cumulative KL preserves lookahead value across zero-marginal-gain context coordinates where single-step $\psi_{\mathrm{Acc}}$ or $\psi_{\mathrm{AUROC}}$ gains vanish.

\subsection{Proof of Theorem~\ref{thm:mnar}, Corollary~\ref{cor:mcar_decoy}, and Proposition~\ref{prop:estimands}}
\begin{proof}
For any $u, \bar{p}, p \in \Delta^{C-1}$ using the \emph{same} measurable selection $s \in \partial \psi$, algebraic expansion yields the generalized three-point Bregman identity \citep{csiszar1991why,banerjee2005clustering}:
\[
  D_\psi^s(u \,\|\, p) - D_\psi^s(u \,\|\, \bar{p}) - D_\psi^s(\bar{p} \,\|\, p) = \langle s(\bar{p}) - s(p),\, u - \bar{p} \rangle.
\]
Setting $u = p(X_{O\cup\{g\}})$, $\bar{p} = \bar{p}_g(x_O)$, and $p = p(x_O)$ and taking $\mathbb{E}[\cdot \mid X_O = x_O, M_g = 1]$ annihilates the cross term because $s(\bar{p}_g(x_O)) - s(p(x_O))$ is $x_O$-measurable and $\mathbb{E}[u - \bar{p}_g(x_O) \mid x_O, M_g=1] = 0$.

\textbf{Corollary~\ref{cor:mcar_decoy}:} Under MCAR ($M_g \perp (X_g, Y) \mid X_O$), $\bar{p}_g(x_O) = \mathbb{E}[p(X_{O\cup\{g\}}) \mid x_O] = p(x_O)$ a.s., so $D_\psi^s(\bar{p}_g \| p) = 0$. Conversely, if $X_g \perp Y \mid X_O, M_g=1$, a Bayes-optimal predictor satisfies $p(X_{O\cup\{g\}}) = \bar{p}_g(x_O)$ a.s.\ on $\{M_g=1\}$, yielding $u_{\mathrm{deconf}}(x_O, g) = 0$ identically. (Scope limit: if $Z$ modifies $P(Y \mid X_O, X_g, Z)$ for an informative test, Theorem~\ref{thm:mnar} strips the additive mask jump $D_\psi^s(\bar{p}_g \| p)$ while the residual remains conditioned on $\{M_g=1\}$.)

\textbf{Proposition~\ref{prop:estimands}:} Cross-entropy is strictly proper \citep{gneiting2007strictly}, so minimizing $\mathbb{E}[\sum_{g \notin O} \mathbf{1}_{\{M_g=1\}} H(p(X_{O\cup\{g\}}), q_\theta(x_O, g))]$ uniquely recovers $\bar{p}_g(x_O)$ on $\{x_O : \mathbb{P}(M_g=1 \mid x_O) > 0\}$. Masking MSE strictly to $(1-a_g)M_g = 1$ (\texttt{elig\_cur}) solves $\min_f \mathbb{E}[(D_\psi^s(p^+ \| \bar{p}_g)/c_g - f(x_O, g))^2 \mid M_g=1]$, giving $r_{\mathrm{cond}}^\star$; regressing $\mathbf{1}_{\{M_g=1\}} D_\psi^s(p^+ \| \bar{p}_g)/c_g$ over all $g \notin O$ (\texttt{unacq}) yields $\mathbb{P}(M_g=1 \mid x_O)\,r_{\mathrm{cond}}^\star(x_O, g)$, which attenuates high-value, selectively ordered tests.
\end{proof}

\subsection{Proof of Theorem~\ref{thm:bellman} (Trapezoidal Area-Bellman \& $L_\infty$ Bound)}
\begin{proof}
Telescoping $m_u - m_0 = \sum_{s=1}^u r_s$ in $\mathrm{AUC}_{\mathrm{trap}} = \frac{1}{k}(\frac{1}{2}m_0 + \sum_{u=1}^{k-1} m_u + \frac{1}{2}m_k) - m_0$ collects coefficient $\frac{k - s + 0.5}{k}$ on $r_s$. For $k=1$, $\mathrm{AUC}_{\mathrm{trap}} = 0.5\, r_1 = \frac{1-0.5}{1}r_1$. For $k \ge 2$, factoring $\frac{k-u+0.5}{k} = \frac{k-1}{k} \cdot \frac{(k-1)-(u-1)+0.5}{k-1}$ for $u \ge 2$ yields $V_k^\star(s) = \max_{g \notin O}\{\frac{k-0.5}{k} r(s,g) + \frac{k-1}{k}\mathbb{E}[V_{k-1}^\star(s') \mid s, g]\}$ by induction.

For the $L_\infty$ error bound, let $e_j = \|\hat{Q}_j - Q_j^\star\|_\infty$. From $e_j \le \varepsilon + \frac{j-1}{j} e_{j-1}$, multiplying by $j$ gives $j e_j - (j-1)e_{j-1} \le j\varepsilon$, summing to $k e_k \le \frac{k(k+1)}{2}\varepsilon \implies e_k \le \frac{k+1}{2}\varepsilon$. Since $Q_j^\star(s, g^\star) - Q_j^\star(s, \hat{g}) \le 2 e_j$, policy suboptimality $\Delta_j = \|V_j^\star - V_j^{\hat{\pi}}\|_\infty$ satisfies $\Delta_j \le 2 e_j + \frac{j-1}{j}\Delta_{j-1} \le (j+1)\varepsilon + \frac{j-1}{j}\Delta_{j-1}$. Multiplying by $j$ yields $j\Delta_j - (j-1)\Delta_{j-1} \le j(j+1)\varepsilon$, summing over $j=1,\dots,B$ to $B \Delta_B \le \frac{B(B+1)(B+2)}{3}\varepsilon$.
\end{proof}

\subsection{Proof of Theorem~\ref{thm:knapsack} and Remark~\ref{rem:mckp} MCKP Counterexample}
\begin{proof}
Ordering eligible pairs by descending ratio $r_1 \ge r_2 \ge \dots \ge r_M$ ($r_e = u_e/c_e$) up to the largest prefix $C_k = \sum_{j=1}^k c_j \le BN$ leaves residual $R = BN - C_k < c_{k+1} \le c_{\max}$ with dual price $\lambda^\star = r_{k+1}$. By Dantzig LP optimality \citep{dantzig1957discrete}, $\mathrm{OPT}_{\mathrm{LP}} = \mathrm{ALG}_{\mathrm{int}} + \lambda^\star R \ge \lambda^\star BN$, giving $\mathrm{OPT}_{\mathrm{LP}} - \mathrm{ALG}_{\mathrm{int}} \le \lambda^\star c_{\max} \le \frac{c_{\max}}{BN}\mathrm{OPT}_{\mathrm{LP}}$.

\textbf{MCKP Heterogeneous-Cost Counterexample:} Consider $N$ identical patients each selecting $\le 1$ panel from $A$ ($u_A = 1.10, c_A = 1.00$, ratio $1.10$) or $B$ ($u_B = 1.50, c_B = 1.50$, ratio $1.00$) under $B_{\mathrm{cap}} = 1.25N$. The true MCKP LP optimum \citep{sinha1979mckp} assigns 50\% to $A$ and 50\% to $B$ using incremental slope $(1.50-1.10)/(1.50-1.00)=0.80$ for total value $1.30N$, whereas greedy $u/c$ puts 100\% on $A$, hits $\sum_g a_{i,g} \le 1$ at $1.00N$, leaves $0.25N$ unspent, and achieves $1.10N$ (15.4\% gap).
\end{proof}

% =====================================================================
% APPENDIX B: IMPLEMENTATION DETAILS
% =====================================================================
\section{Architecture Details and Code-to-Theory Matrix}
\label{app:impl}

All experiments were implemented in Python~3.12 with PyTorch~2.7, scikit-learn, LightGBM, XGBoost, and SciPy. \texttt{DualScaleCrossResPredictor} uses token dimension $d_{\mathrm{tok}}=12$, hidden width $d_h=256$, two \texttt{LowRankCrossLN} layers ($r_1=48, r_2=32$), and 3 \texttt{PreNormResBlock} layers (dropout $0.08$), trained with Adam, cosine annealing, and subset masks sampled from $\{1/6 \text{ empty}, 1/3 \text{ singleton}, 1/2 \text{ uniform}\}$. \texttt{LIBRAPolicyNet} passes state summary $[x_{\mathrm{raw}}(O); h_{\mathrm{feat}}(O); p_\phi(x_O); a_O; k/K_{\max}]$ through a 3-block pre-norm residual MLP ($d_\pi=256$) with heads \texttt{head\_pbar}, \texttt{head\_u1}, and \texttt{head\_qk} ($\tau_{\mathrm{EMA}} = 0.02$). LightGBM uses up to 700 trees, lr $0.04$, 47 leaves, min\_child\_samples $40$, subsample $0.85$, colsample $0.80$, $L_1=1.0, L_2=2.0$; XGBoost uses histogram trees, depth $6$, min\_child\_weight $20$.

\begin{table}[h]
  \caption{Implementation reconciliation matrix across experimental suites 105--112.}
  \label{tab:app_matrix}
  \centering
  \scriptsize
  \begin{tabular}{@{}lcccccc@{}}
    \toprule
    Mechanism & Exp 105 & Exp 106 & Exp 107 & Exp 108 & Exp 109 / 111 & Exp 110 / 112 \\
    \midrule
    \texttt{head\_pbar} MNAR deconf. & Yes & Yes & Yes & Optional & Yes & Yes \\
    MSE loss mask & \texttt{unacq} & \texttt{elig\_cur} & \texttt{elig\_cur} & \texttt{unacq} & \texttt{unacq} & \texttt{elig\_cur} (A/B) \\
    $k_{\mathrm{rem}}=2$ backup & $0.5\, u_1$ & $0.5\, u_1$ & $Q_k$ & $Q_k$ & $Q_k$ & $Q_k$ \\
    EMA Double-Q & No & No & No & No & \textbf{Yes} & On-Policy / Yes \\
    Val.\ temp.\ calib. & Yes & Post-hoc & Post-hoc & Yes & Yes & Post-hoc \\
    \bottomrule
  \end{tabular}
\end{table}

% =====================================================================
% APPENDIX C: PUBLIC BENCHMARKS
% =====================================================================
\section{Extended Public Blind Benchmark Results}
\label{app:public}

\begin{figure}[h]
  \centering
  \includegraphics[width=\textwidth]{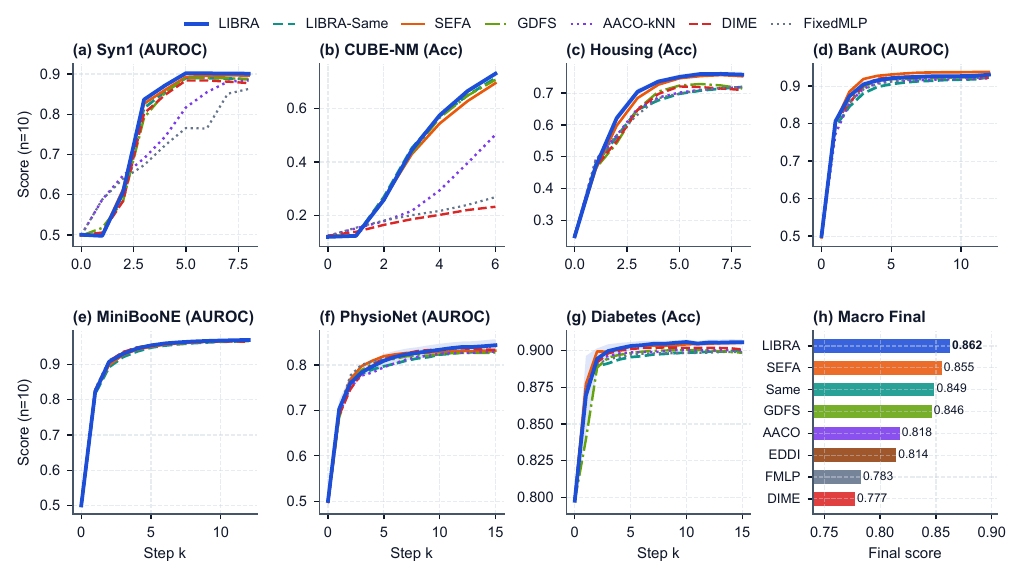}
  \caption{\textbf{All seven public benchmark sequential acquisition curves ($N=70$ blind runs, Seeds 5--14).} \textbf{(a--g)} Mean accuracy or AUROC across steps $k$ ($n=10$ blind seeds per dataset; shaded band $\pm 1$ s.d. for LIBRA). \textbf{(h)} Macro full-feature terminal score across all 8 methods.}
  \label{fig:app_pub}
\end{figure}

\begin{table}[h]
  \caption{Pre-seal validation Double-Q selection (Seeds 0--2, Exp.~109) and standalone blind batch summaries ($N=35$ each).}
  \label{tab:app_val}
  \centering
  \scriptsize
  \resizebox{\columnwidth}{!}{\begin{tabular}{@{}lcccccc@{}}
    \toprule
    Batch / Config & Method & Macro Seq-AUC & Macro Q50 & Macro Final & Seq wins (Holm $t$-$P$) & Q50 wins (Holm $t$-$P$) \\
    \midrule
    Val (Seeds 0--2) & \texttt{Double-Q=False / True} & 0.7237 / \textbf{0.7244} & 0.5880 / \textbf{0.5895} & 0.8500 / \textbf{0.8511} & --- & --- \\
    \midrule
    Batch 1 (Seeds 5--9) & LIBRA\_v11 vs.\ SEFA & \textbf{0.7685} vs.\ 0.7640 & \textbf{0.6597} vs.\ 0.6412 & \textbf{0.8646} vs.\ 0.8553 & 24/35 (0.0058) & 35/35 ($5.58\times 10^{-11}$) \\
    Batch 2 (Seeds 10--14) & LIBRA\_v11 vs.\ SEFA & \textbf{0.7656} vs.\ 0.7634 & \textbf{0.6597} vs.\ 0.6414 & \textbf{0.8604} vs.\ 0.8552 & 23/35 (0.1405) & 33/35 ($1.03\times 10^{-8}$) \\
    \bottomrule
  \end{tabular}}
\end{table}

\begin{table}[h]
  \caption{Per-dataset pooled 10-seed Seq-AUC / Q50 across seven public benchmarks (Seeds 5--14).}
  \label{tab:app_perds}
  \centering
  \scriptsize
  \setlength{\tabcolsep}{3.5pt}
  \resizebox{\textwidth}{!}{\begin{tabular}{@{}lcccccccc@{}}
    \toprule
    Dataset (Metric) & LIBRA & SamePred & SEFA & GDFS & AACO-kNN & DIME & FixedMLP & EDDI-CMI \\
    \midrule
    Syn1 (AUROC) & \textbf{0.777} / \textbf{0.658} & 0.767 / 0.650 & 0.773 / 0.635 & 0.764 / 0.586 & 0.739 / 0.648 & 0.760 / 0.598 & 0.711 / 0.604 & 0.737 / 0.646 \\
    CUBE-NM (Acc) & \textbf{0.415} / \textbf{0.185} & 0.414 / 0.183 & 0.399 / 0.169 & 0.409 / 0.158 & 0.259 / 0.182 & 0.183 / 0.160 & 0.199 / 0.177 & 0.255 / 0.181 \\
    Housing (Acc) & \textbf{0.663} / \textbf{0.526} & 0.621 / 0.501 & 0.654 / 0.494 & 0.627 / 0.434 & 0.626 / 0.493 & 0.625 / 0.451 & 0.623 / 0.497 & 0.603 / 0.464 \\
    Bank (AUROC) & 0.891 / \textbf{0.797} & 0.877 / 0.790 & \textbf{0.900} / 0.791 & 0.887 / 0.757 & 0.880 / 0.780 & 0.889 / 0.767 & 0.884 / 0.773 & 0.866 / 0.767 \\
    MiniBooNE (AUROC) & \textbf{0.924} / \textbf{0.881} & 0.918 / 0.854 & 0.921 / 0.857 & \textbf{0.924} / 0.771 & \textbf{0.924} / 0.860 & \textbf{0.924} / 0.814 & \textbf{0.924} / 0.847 & 0.881 / 0.815 \\
    PhysioNet (AUROC) & 0.801 / \textbf{0.707} & 0.792 / 0.692 & 0.801 / 0.686 & 0.794 / 0.660 & 0.790 / 0.691 & 0.799 / 0.671 & \textbf{0.804} / 0.680 & 0.790 / 0.678 \\
    Diabetes (Acc) & \textbf{0.898} / \textbf{0.864} & 0.892 / 0.857 & \textbf{0.898} / 0.858 & 0.891 / 0.796 & 0.894 / 0.848 & 0.895 / 0.835 & 0.892 / 0.853 & 0.889 / 0.836 \\
    \bottomrule
  \end{tabular}}
\end{table}

\FloatBarrier
% =====================================================================
% APPENDIX D: MNAR TABLES
% =====================================================================
\section{Controlled Causal MNAR Intervention Tables}
\label{app:mnar}

\begin{table}[h]
  \caption{Controlled Synthetic EHR ($\alpha = 3.0$) and Semi-Synthetic PhysioNet ($\alpha = 3.47$) MNAR interventions (Exp.~108 \& 110C).}
  \label{tab:app_mnar}
  \centering
  \scriptsize
  \resizebox{\textwidth}{!}{\begin{tabular}{@{}lcccccc@{}}
    \toprule
    Benchmark / Setting & Method & World A Score & World B Seq/Final & World B Q50 & Step-1 Diag (\%) & Q50 / All Decoy (\%) \\
    \midrule
    Synth $\alpha=3.0$ (e2e) & FixedMLP / DIME / GDFS & 0.771 / 0.726 / 0.742 & 0.760 / 0.672 / 0.709 & 0.686 / 0.641 / 0.650 & --- & 21.5 / 46.9 / 58.7 (Q50) \\
    Synth $\alpha=3.0$ (e2e) & SEFA / LIBRA-NoDeconf & 0.775 / 0.780 & 0.761 / 0.771 & 0.681 / 0.694 & --- & 16.2 / 9.09 (Q50) \\
    Synth $\alpha=3.0$ (e2e) & \textbf{LIBRA-Deconfounded} & 0.762 & \textbf{0.8101} & \textbf{0.6990} & --- & \textbf{0.00} (Q50) / \textbf{0.04} (All) \\
    \midrule
    PhysioNet $\alpha=3.47$ (e2e) & FixedMLP / DIME / NoDeconf & 0.943 / 0.938 / 0.923 & 0.742 / 0.721 / 0.773 & 0.597 / 0.603 / 0.606 & --- & 100.0 / 96.0 / 100.0 (Q50) \\
    PhysioNet $\alpha=3.47$ (e2e) & SEFA / AACO-kNN / \textbf{LIBRA} & 0.886 / 0.820 / 0.808 & 0.783 / \textbf{0.803} / 0.796 & 0.655 / 0.679 / \textbf{0.689} & --- & 26.5 / 0.08 / \textbf{0.00} (Q50) \\
    \midrule
    PhysioNet (Same \texttt{obs\_pred}) & DIME / NoDeconf & 0.931 / 0.915 & 0.707 / 0.779 & 0.593 / 0.609 & 0.0 / 59.6\% & 13.3 / 13.3 (All) \\
    PhysioNet (Same \texttt{obs\_pred}) & \textbf{LIBRA-Deconf / OnPolicy} & 0.806 / 0.812 & 0.795 / \textbf{0.796} & \textbf{0.704} / 0.700 & \textbf{100.0 / 100.0\%} & 6.28 / \textbf{6.01} (All) \\
    \bottomrule
  \end{tabular}}
\end{table}

\FloatBarrier
% =====================================================================
% APPENDIX E: CLINICAL COHORTS, TRANSFER & EXTENDED FIGURE
% =====================================================================
\section{Clinical Cohorts, Transfer, and Extended Clinical Results}
\label{app:clinical}

\begin{table}[h]
  \caption{Transnational critical-care cohorts (77 hospitals, 75,482 stays, 208,686 total windows).}
  \label{tab:app_cohorts}
  \centering
  \scriptsize
  \resizebox{\textwidth}{!}{\begin{tabular}{@{}lccccccc@{}}
    \toprule
    Cohort & Holdout Design & Total Stays (Win.) & Test Stays (Win.) & Train/Test Rate & Free $d_0$ & Panels (mL) & Routine Draw \\
    \midrule
    \textbf{eICU Multicenter} & 61 dev $\to$ 14 unseen US hosp. & 30,810 (83,573) & 5,788 (15,374) & 34.5\% / 26.8\% & 103 & 4 (1.5--3.5) & 5.43\,mL \\
    \textbf{MIMIC-IV CVICU} & Boston patient-disjoint & 14,802 (43,080) & 2,225 (6,464) & 34.1\% / 36.1\% & 103 & 4 (1.5--3.5) & 7.51\,mL \\
    \textbf{Fuwai Decomp.} & Beijing 2022 $\to$ 2023 & 29,870 (78,215) & 17,991 (45,781) & 27.3\% / 20.7\% & 152 & 5 (1.5--3.5) & 8.52\,mL \\
    \textbf{Fuwai Incident AKI} & Beijing 2022 $\to$ 2023 (6--72\,h) & 3,818 (3,818) & 2,141 (2,141) & 21.9\% / 15.7\% & 152 & 5 (1.5--3.5) & 5.11\,mL \\
    \bottomrule
  \end{tabular}}
\end{table}

\begin{table}[h]
  \caption{Tri-cohort single-seed mean AUROC ($n=5$) and 5-seed calibrated ensemble AUROC / AUPRC at $B=1.00$ (Exp.~106).}
  \label{tab:app_clin9}
  \centering
  \scriptsize
  \setlength{\tabcolsep}{3.5pt}
  \resizebox{\textwidth}{!}{\begin{tabular}{@{}lcccc@{}}
    \toprule
    Method ($B=1.00$) & eICU (Single / Ens.) & MIMIC-IV (Single / Ens.) & Fuwai Decomp.\ (Single / Ens.) & Fuwai AKI (Single / Ens.) \\
    \midrule
    Routine BloodGas & 0.7160 / 0.7371 & 0.7695 / 0.7912 & 0.7437 / 0.7645 & 0.7811 / 0.8199 \\
    Routine ChemRenal & 0.7058 / 0.7299 & 0.7640 / 0.7854 & 0.7450 / 0.7650 & 0.7817 / 0.8204 \\
    FixedMLP & 0.7171 / 0.7388 & 0.7708 / 0.7912 & 0.7459 / 0.7649 & 0.7808 / 0.8195 \\
    EDDI-CMI & 0.7388 / 0.7526 & 0.7860 / 0.7982 & 0.7561 / 0.7652 & 0.8090 / \textbf{0.8308} \\
    GDFS & 0.7190 / 0.7382 & 0.7861 / 0.8060 & 0.7521 / 0.7672 & 0.7797 / 0.8113 \\
    DIME & 0.7190 / 0.7452 & 0.7677 / 0.7974 & 0.7493 / 0.7652 & 0.7623 / 0.8029 \\
    AACO-kNN & 0.7346 / 0.7541 & 0.7812 / 0.8024 & 0.7468 / 0.7669 & 0.7827 / 0.8193 \\
    SEFA & 0.7289 / 0.7526 & 0.7627 / 0.7978 & 0.7365 / 0.7654 & 0.7615 / 0.7929 \\
    \textbf{LIBRA (Ours)} & \textbf{0.7632} / \textbf{0.7686} & \textbf{0.8059} / \textbf{0.8112} & \textbf{0.7626} / \textbf{0.7673} & \textbf{0.8145} / 0.8234 \\
    \bottomrule
  \end{tabular}}
\end{table}

\begin{table}[h]
  \caption{Bidirectional zero-label transfer (MIMIC-IV $\leftrightarrow$ eICU, $B=1.00$) and \texttt{UnlabeledTargetStd} ablation (Exps.~107B \& 110B).}
  \label{tab:app_xfer}
  \centering
  \scriptsize
  \resizebox{\textwidth}{!}{\begin{tabular}{@{}lcccc@{}}
    \toprule
    Method & MIMIC$\to$eICU Ens.\ AUROC & \texttt{NoStd$\to$UnlabStd} ($\Delta$, $P$) & eICU$\to$MIMIC Ens.\ AUROC & \texttt{NoStd$\to$UnlabStd} ($\Delta$, $P$) \\
    \midrule
    \textbf{LIBRA\_Policy\_SameStrong} & \textbf{0.6793} & $0.6689\to 0.6725$ ($+0.35$, 0.180) & \textbf{0.7494} & $0.7280\to 0.7437$ ($+1.57$, \textbf{0.010}) \\
    \textbf{LIBRA (end-to-end)} & 0.6783 & $0.6492\to 0.6636$ ($+1.44$, \textbf{0.010}) & 0.7484 & $0.7272\to 0.7439$ ($+1.67$, \textbf{0.008}) \\
    AACO\_kNN\_SameStrong & 0.6783 & --- & 0.7365 & --- \\
    GDFS / DIME & 0.6460 / 0.6643 & --- & 0.7391 / 0.7230 & --- \\
    SEFA & 0.6284 & $0.5912\to 0.6157$ ($+2.45$, 0.019) & 0.7260 & $0.6965\to 0.6882$ ($-0.83$, 0.564) \\
    \bottomrule
  \end{tabular}}
\end{table}

\begin{table}[h]
  \caption{Iso-accuracy blood savings vs.\ $B=2.00$ comparators (Exp.~112): LIBRA achieves $\ge$ comparator AUROC at 70.5--100\% less blood.}
  \label{tab:app_iso}
  \centering
  \scriptsize
  \resizebox{\textwidth}{!}{\begin{tabular}{@{}lcccc@{}}
    \toprule
    Cohort & LIBRA $B=0.50$ (AUROC; mL) & vs.\ Routine BG ($B=2.0$) & vs.\ SEFA ($B=2.0$) & vs.\ AACO-kNN ($B=2.0$) \\
    \midrule
    eICU (14 unseen) & 0.7534 (1.11\,mL) & 100\% / 73.1\% interp.\ saved & 73.1\% / 77.7\% interp. & 72.1\% / 79.5\% interp. \\
    MIMIC-IV CVICU & 0.7946 (0.83\,mL) & 75.6\% / 81.2\% interp. & 100\% / 100\% interp. & 79.1\% / 85.4\% interp. \\
    Fuwai Decomp. & 0.7605 (0.80\,mL) & 100\% saved (beats at $B=0$) & 100\% saved & 100\% saved \\
    \bottomrule
  \end{tabular}}
\end{table}

\begin{figure}[h]
  \centering
  \includegraphics[width=\textwidth]{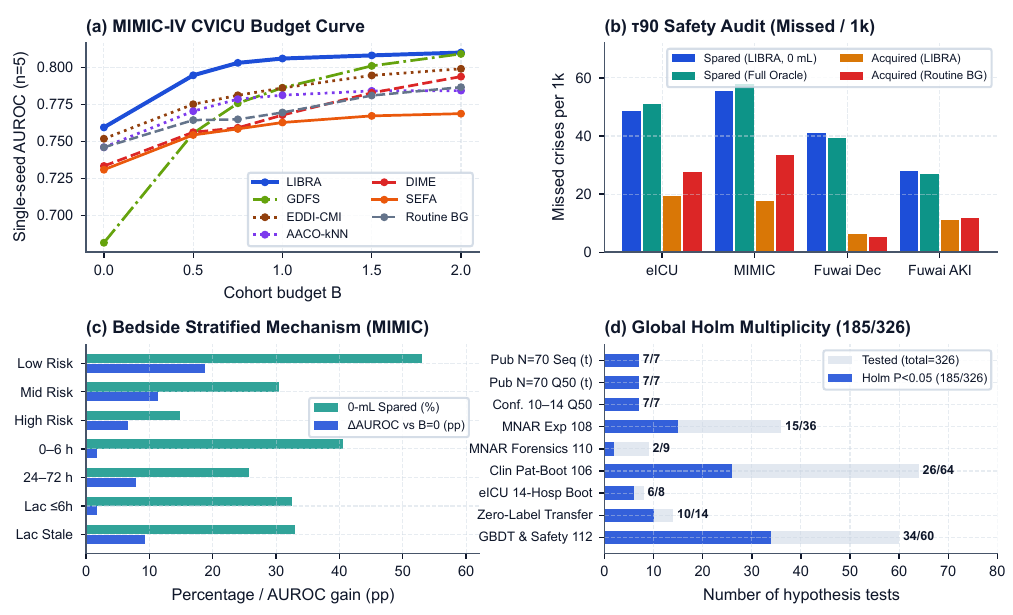}
  \caption{\textbf{Extended clinical validation.} \textbf{(a)} MIMIC-IV CVICU budget curves. \textbf{(b)} Dual-stratum $\tau_{90}$ missed-event safety audit across all four cohorts ($B=1.00$). \textbf{(c)} Bedside stratified routing on MIMIC-IV (0-mL spared \% and $\Delta$AUROC over $B=0$). \textbf{(d)} Global Holm--Bonferroni multiplicity summary across 326 pre-specified tests in 18 families (185/326 significant).}
  \label{fig:app_clin}
\end{figure}

\FloatBarrier
% =====================================================================
% APPENDIX F: CRYPTOGRAPHIC SEALS
% =====================================================================
\section{Cryptographic Seals and Reproducibility Audit}
\label{app:seals}

All experimental suites (Experiments 105--112) were pre-sealed with SHA-256 source hashes (\texttt{SEAL105.json}--\texttt{SEAL112.json}) and independently replayed via deterministic arithmetic verification scripts (\texttt{verify105.py}--\texttt{verify112.py}), confirming zero patient-level exports (\texttt{privacy\_zero\_patient\_files = true}) and floating-point summary agreement $\le 1.42 \times 10^{-14}$ across all checks.

\end{document}